\documentclass[10pt,twocolumn,letterpaper]{article}

\usepackage{iccv}              

\usepackage{amssymb}
\usepackage{pifont}
\newcommand{\cmark}{\ding{51}}
\newcommand{\xmark}{\ding{55}}
\usepackage{amsmath, graphicx, booktabs}

\makeatletter
\newcommand{\labeltext}[2]{%
  \@bsphack
  \MakeLinkTarget*{#1}%
  \def\@currentlabel{#1}{\label{#2}}%
  \@esphack
}
\makeatother

\definecolor{iccvblue}{rgb}{0.21,0.49,0.74}
\usepackage[pagebackref,breaklinks,colorlinks,allcolors=iccvblue]{hyperref}

\def\paperID{14016} 
\def\confName{ICCV}
\def\confYear{2025}

\title{A Multimodal Physics-Informed Neural Network Approach for Mean Radiant Temperature Modeling}

\author{
Pouya Shaeri\\
School of Computing and Augmented Intelligence, Arizona State University\\
{\tt\small pshaeri@asu.edu}
\and
Saud AlKhaled\\
College of Architecture, Kuwait University\\
{\tt\small salkhaled@ku.edu.kw}
\and
Ariane Middel\\
School of Arts, Media and Engineering, Arizona State University\\
{\tt\small ariane.middel@asu.edu}
}

\begin{document}
\maketitle
\begin{abstract}
Outdoor thermal comfort is a critical determinant of urban livability, particularly in hot desert climates where extreme heat poses challenges to public health, energy consumption, and urban planning. Mean Radiant Temperature ($T_{mrt}$) is a key parameter for evaluating outdoor thermal comfort, especially in urban environments where radiation dynamics significantly impact human thermal exposure. Traditional methods of estimating $T_{mrt}$ rely on field measurements and computational simulations, both of which are resource intensive. This study introduces a Physics-Informed Neural Network (PINN) approach that integrates shortwave and longwave radiation modeling with deep learning techniques. By leveraging a multimodal dataset that includes meteorological data, built environment characteristics, and fisheye image-derived shading information, our model enhances predictive accuracy while maintaining physical consistency. Our experimental results demonstrate that the proposed PINN framework outperforms conventional deep learning models, with the best-performing configurations achieving an RMSE of 3.50 and an $R^2$ of 0.88. This approach highlights the potential of physics-informed machine learning in bridging the gap between computational modeling and real-world applications, offering a scalable and interpretable solution for urban thermal comfort assessments.
\end{abstract}

\noindent \textbf{Keywords:} Mean Radiant Temperature, Physics-Informed Neural Network, Outdoor Thermal Comfort, Urban Climate, Multimodal Deep Learning, Computer Vision Radiation Modeling.    
\section{Introduction}
\label{sec:introduction}

\begin{figure*}[!t]
    \centering
    \includegraphics[width=0.9\linewidth]{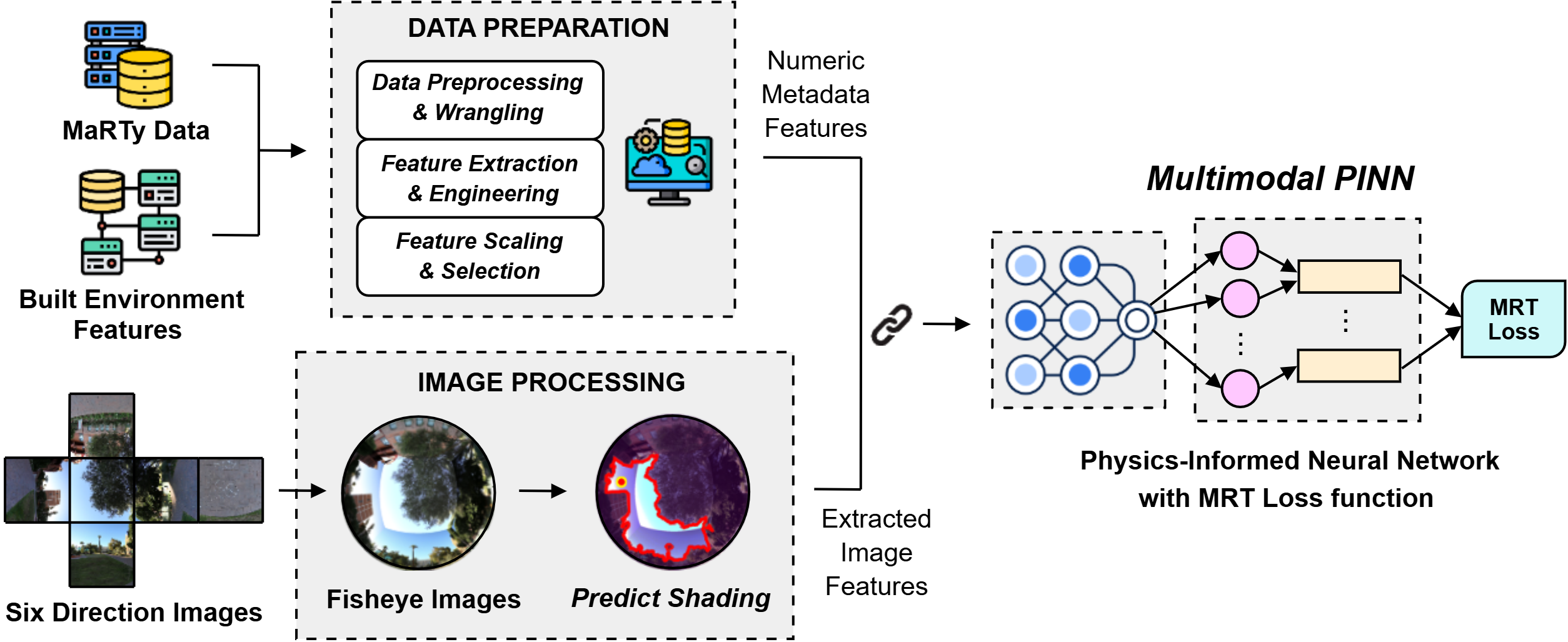}
    \caption{Overview of the proposed Physics-Informed Neural Network (PINN) workflow for $T_{mrt}$ estimation.}
    \label{fig:workflow}
\end{figure*}

Climate change and rapid urbanization are intensifying heat exposure in cities, exacerbating urban heat islands and increasing the risk of heat-related health issues, energy consumption, and infrastructure stress \cite{vanos2023physiological, li2017surface, guzman2025heatsuite}. One of the most important metrics for evaluating thermal exposure in outdoor environments is Mean Radiant Temperature ($T_{mrt}$), which quantifies the combined effect of shortwave and longwave radiation on a human body from surrounding surfaces and atmospheric radiation sources \cite{middel2014impact, crank2020validation}. In urban settings, where complex interactions between solar radiation, surface reflectance, shading from buildings and vegetation, and thermal emissions from materials significantly affect microclimate conditions, accurate $T_{mrt}$ estimation is crucial for designing sustainable cities and improving public health outcomes \cite{lindberg2020urban, alkhaled2024webmrt}.

Traditional approaches to estimating $T_{mrt}$ generally fall into two categories: direct field measurements and computational simulations. Field measurements rely on instruments such as black globe thermometers, radiometers, and six-directional radiation sensors to capture radiative fluxes with high accuracy \cite{meggers2022improving, middel2019micrometeorological}. Despite their reliability, these methods are impractical for large-scale applications due to their dependence on costly and labor-intensive equipment, making long-term or real-time monitoring considerably more difficult~\cite{eldesoky2021high, azghan2024cudle}. On the other hand, computational simulations based on radiative transfer equations, such as those implemented in models like ENVI-met and SOLWEIG, simulate shortwave and longwave radiation fluxes based on urban geometry, atmospheric properties, and material emissivity \cite{crank2020validation, lindberg2020urban}. 
While these models provide physically consistent results, they require extensive parameter tuning, high computational resources, and detailed site-specific data, which limit their adaptability and scalability to dynamic, large-scale urban environments.

Recent advancements in deep learning have opened new avenues for estimating $T_{mrt}$ using machine learning models trained on meteorological and environmental data \cite{alkhaled2024webmrt, xie2022artificial}. Neural networks have demonstrated the ability to learn complex spatial and temporal patterns of radiation exposure, allowing for faster and potentially scalable $T_{mrt}$ predictions. However, data-driven models often suffer from overfitting, lack physical interpretability, and fail to generalize effectively across diverse urban morphologies and climatic conditions \cite{jia2024estimation}. 

To address these challenges, we propose a multimodal Physics-Informed Neural Network (PINN) framework (Figure \ref{fig:workflow}) that combines deep learning techniques with fundamental radiative transfer principles to enhance the accuracy, interpretability, and generalizability of $T_{mrt}$ estimation, leveraging the advanced representations that deep neural networks achieve through multimodal learning~\cite{qu2024deep, beigi2024lrq}. Unlike conventional machine learning approaches, PINNs incorporate governing physical equations directly into the neural network training process, ensuring that predictions adhere to thermodynamic constraints \cite{raissi2019physics, cai2021physics, afzal2025hyperparameter}. Our method enhances predictive performance by incorporating the six-directional $T_{mrt}$ formula into the loss function while leveraging multimodal data sources, including numeric measured environmental features (air temperature, humidity, wind speed, solar angles) and fisheye image-derived features (capturing the shading conditions and surrounding built environment features). 

Shading and solar exposure play a pivotal role in outdoor thermal comfort, particularly in densely built urban areas where sky obstruction varies significantly~\cite{middel202150, middel2016impact}. Traditional $T_{mrt}$ models struggle to capture the spatial heterogeneity of shading effects due to the complexity of urban structures. By combining physics-informed deep learning with vision-based urban climate modeling, our approach significantly improves the ability to estimate $T_{mrt}$ in heterogeneous urban environments.

\vspace{0.1cm}
\noindent
\textbf{Research Questions}\\
This study aims to address the following research questions:

\noindent \textbf{RQ1}\labeltext{RQ1}{rq1}: How does a Physics-Informed Neural Network (PINN) compare to traditional machine learning and empirical methods in terms of accuracy and physical consistency for $T_{mrt}$ estimation?

\noindent \textbf{RQ2}\labeltext{RQ2}{rq2}: How does the incorporation of fisheye image-derived shading information impact the predictive performance of $T_{mrt}$ models?

\noindent \textbf{RQ3}\labeltext{RQ3}{rq3}: How generalizable is the multimodal deep learning framework in integrating metadata, solar position parameters, and computer vision-derived shading features across diverse urban settings?

\noindent \textbf{RQ4}\labeltext{RQ4}{rq4}: What are the key physical constraints that should be embedded in the neural network loss function to ensure thermodynamic consistency in $T_{mrt}$ predictions?

\noindent \textbf{RQ5}\labeltext{RQ5}{rq5}: To what extent can our multimodal deep learning framework reduce the need for extensive sensor-based measurements while maintaining accurate $T_{mrt}$ estimation?

\noindent
\textbf{Our Contribution}\\
\noindent
Previous methods for $T_{mrt}$ estimation have relied on measured field data or simulation-based approaches. We propose a framework that reduces reliance on expensive sensors and field measurements by developing a validated model capable of maintaining high accuracy even when certain input features are removed. Using a PINN, our approach ensures thermodynamic consistency while improving generalizability across different environments. Additionally, we leverage computer vision techniques, specifically fisheye imagery, to dynamically extract spatial shading features, effectively capturing crucial environmental factors that influence radiative flux. Through intelligent feature selection and multimodal integration, our model serves as a proxy for traditional methods, providing an adaptable and physically consistent solution for urban thermal comfort assessment without the need for extensive instrumentation.
\section{Related Work}
\label{sec:related_work}

Calculating the Mean Radiant Temperature ($T_{mrt}$) has been a subject of extensive research in human biometeorology and urban climate, with various methods developed to improve accuracy in heterogeneous built environments. Traditional approaches relied primarily on physical measurements and empirical models, while more recent studies have leveraged machine learning techniques to enhance predictive performance. However, despite these advances, many existing methods have notable limitations in their adaptability, computational efficiency, and robustness when applied to complex outdoor settings with dynamic shading conditions. In this section, we review prior studies in three key areas: Urban Climate and Heat Assessment, Physics-Informed Neural Networks (PINNs), and Computer Vision for Radiation Estimation. We also discuss the MaRTy dataset, which forms the basis for our study.

\subsection{Urban Climate and Heat Assessment}

Accurate estimation of $T_{mrt}$ is crucial for urban climate research, sustainable city planning, and the development of thermally comfortable public spaces. Traditional methods for estimating $T_{mrt}$ generally fall into two categories. Field measurements rely on instruments such as black globe thermometers, radiometers, and six-directional radiation sensors to provide direct measurements of thermal radiation~\cite{meggers2022improving,middel2019micrometeorological}. While highly accurate, these methods require extensive instrumentation, making large-scale, long-term, or real-time applications impractical. Computational simulations based on radiative transfer equations, such as those used in ENVI-met and SOLWEIG, estimate $T_{mrt}$ by simulating shortwave and longwave radiation fluxes. Despite their precision, these models are computationally expensive and heavily dependent on detailed input parameters, often requiring extensive calibration and site-specific data~\cite{sinsel2022implementation}.

Recent studies have investigated machine learning approaches for $T_{mrt}$ estimation. Jia et al.~\cite{jia2024estimation} compared five different $T_{mrt}$ estimation techniques, including deep neural networks (DNNs), showing that DNNs outperformed traditional methods. However, their approach focused on statistical learning without incorporating physical constraints or multimodal data sources, making it less interpretable and more sensitive to dataset biases. Xie et al.~\cite{xie2022backpropagation} introduced a genetic algorithm-optimized backpropagation neural network (GA-BPNN) for long-term $T_{mrt}$ prediction, achieving a mean absolute percentage error (MAPE) below 1\%. Yet, their reliance on meteorological parameters without integrating environmental visual data limited their model's adaptability to urban microclimates.


\subsection{Physics-Informed Neural Networks (PINNs)}

Physics-Informed Neural Networks (PINNs) have emerged as a robust framework for embedding domain knowledge into deep learning models\cite{aghaei2024pinnies, raissi2017physics}. PINNs combine neural network approximations with physics-based constraints to ensure physically consistent predictions~\cite{raissi2019physics, huhn2023physics}. Various studies have demonstrated the effectiveness of PINNs in solving complex differential equations.

Biswal et al.\cite{biswal2024radiation} applied PINNs to solve the Radiative Transfer Equation (RTE), demonstrating the method’s potential to handle computational complexities associated with radiation modeling and validating its accuracy in solving one-dimensional RTE problems. Their findings highlight the broader applicability of PINNs in radiative heat transfer modeling. Firoozsalari et al.\cite{firoozsalari2024machine} applied PINNs to solve Fokker–Planck equations, demonstrating how physics constraints enhance convergence and accuracy. Building on this, they introduced DeepFDENet \cite{firoozsalari2023deepfdenet}, showcasing PINNs' adaptability in solving fractional differential equations and broader applicability in complex physical systems.

Krishnapriyan et al.~\cite{krishnapriyan2021characterizing} explored the effectiveness of PINNs in solving complex physical systems, identifying challenges in convergence and proposing strategies to improve their accuracy in high-dimensional differential equations. Parand et al.~\cite{parand2024neural} applied PINNs to solve nonlinear differential equations of Lane–Emden type, an approach closely related to our work, as we employ PINNs to solve radiative transfer equations governing $T_{mrt}$ estimation. Additionally, other recent works have applied PINNs to thermal modeling in urban and building environments, including thermal cavity flow simulations~\cite{fowler2024physics, aghaei2023solving} and building energy modeling~\cite{chen2023physics}.


\subsection{Computer Vision for Radiation Estimation}

Computer vision has increasingly been applied in thermal environment monitoring and radiation estimation~\cite{clero2023review}. Infrared (IR) imaging technology enables non-contact temperature measurement by detecting thermal radiation emitted by objects, making it particularly useful for urban heat mapping and building thermal assessments~\cite{yang2023computer, kempelis2024computer}. 

Some studies have used thermal IR cameras to directly measure skin temperature variations in human thermal stress analysis~\cite{engert2014exploring}. Other works have explored radiative energy estimation using vision-based approaches, including techniques for high-temperature surface measurements in industrial applications~\cite{fabijanska2009computer}. 

For urban applications, researchers have leveraged fisheye images to calculate the Sky View Factor (SVF), an essential parameter for estimating local shading and radiation fluxes~\cite{middel2018sky}. Our work builds upon previous methodologies by integrating Physics-Informed Neural Networks (PINNs) to enforce thermodynamic consistency while leveraging fisheye imagery to enhance spatial awareness of radiative flux distributions. By extending PINNs to multimodal data fusion, we incorporate environmental metadata and image-derived features, improving the predictive accuracy of $T_{mrt}$ models. This integration enables precise spatial characterization of shading effects, ensuring more reliable $T_{mrt}$ estimations in heterogeneous urban environments.

\subsection{MaRTy Dataset}

We developed our PINN based on the MaRTy dataset ~\cite{middel202150}, providing high-resolution microclimate measurements collected during nine field campaigns (2016–19) at 159 locations in Tempe, Arizona using the MaRTy cart. Developed by Middel et al.~\cite{middel2019micrometeorological}, the MaRTy cart is an advanced mobile weather station designed specifically for urban climate studies. The human-biometeorological instrument platform is equipped with three net radiometers measuring shortwave and longwave flux densities in six directions (up, down, left, right, front, back), along with sensors for air temperature, surface temperature, humidity, wind speed, and GPS position. Image cubes of the surroundings are captured concurrently, enabling the characterization of nearby tree canopies and shading features, making it a uniquely valuable dataset for our model. This robust approach provides precise estimates of Mean Radiant Temperature ($T_{mrt}$) and supports versatile, high-spatiotemporal-resolution microclimate assessments.

The data set consists of multimodal environmental observations, including:
\begin{itemize}
    \item \textbf{Meteorological parameters:} Air temperature, relative humidity, wind speed, date, time, latitude, longitude and altitude.
    \item \textbf{Radiative flux measurements:} Six-directional shortwave and longwave radiation readings from net radiometers.
    \item \textbf{Built environment descriptors:} Surrounding surface materials, tree canopy cover, percentage of surrounding buildings and pervious-impervious surfaces.
    \item \textbf{Solar geometry-derived parameters:} Solar altitude, solar azimuth, and minutes from sunrise.
    \item \textbf{Fisheye image-derived features:} image-extracted features and dynamic shade assessment.
\end{itemize}


\vspace{0.1cm}
\noindent
\textbf{MaRTy in Prior Research.} One of the key studies utilizing MaRTy is the ``Fifty Grades of Shade" study by Middel et al.~\cite{middel202150}, where the impact of urban shading on pedestrian thermal comfort was analyzed. This work demonstrated MaRTy's effectiveness in capturing fine-scale radiation variations across different urban morphologies. Similarly, WebMRT~\cite{alkhaled2024webmrt} leveraged MaRTy data to develop a machine learning-based $T_{mrt}$ estimation model, incorporating solar positioning, surface materials, and meteorological parameters. Our study builds upon these works by integrating physics-informed learning and computer vision techniques into the estimation framework.

\noindent
\textbf{Dataset Summary.} The MaRTy dataset used in this study consists of the attributes in table \ref{tab:marty_summary}.
\begin{table}[htbp]
    \centering
    \small
    \renewcommand{\arraystretch}{0.95} 
    \setlength{\tabcolsep}{3pt} 
    \begin{tabular}{ll}
        \toprule
        \textbf{Attribute} & \textbf{Details} \\
        \midrule
        No. of Observations (with images) & 1,130 \\
        Time Range & 07:30 – 20:30 \\
        Date Rang & June – August, 2018 \\
        Data Types & Numeric, Nominal, Image \\
        Air Temperature ($^\circ$C) & 26.02 – 43.97 \\
        $T_{mrt}$ ($^\circ$C) & 15.54 – 76.22 \\
        \bottomrule
    \end{tabular}
    \caption{Summary of MaRTy dataset observations used in this study.}
    \label{tab:marty_summary}
\end{table}
\vspace{-0.2cm}
\section{Methodology}
\label{sec:methodology}

Our proposed workflow begins with data acquisition and preprocessing using the MaRTy dataset \cite{middel2019micrometeorological}, which includes meteorological data and six-directional images. Preprocessing involved handling missing values, detecting sensor noise, and deriving solar geometry parameters such as solar altitude, azimuth, and minutes from sunrise to track radiative flux variations. We transformed the six-directional images into an upward-view fisheye projection to efficiently capture sky exposure, shading, and surrounding structures. SegFormer~\cite{xie2021segformer} was used for sky masking and shading estimation, while ResNet-50~\cite{koonce2021resnet}, extracted spatial features of the built environment by unfreezing the last 30 layers and fine-tuning. These extracted features, combined with metadata, were integrated into a PINN to predict shortwave and longwave radiation components for $T_{mrt}$ estimation. The PINN framework leveraged the six-directional $T_{mrt}$ equation to enforce thermodynamic consistency, ensuring physically realistic predictions. Model performance was evaluated using RMSE and $R^2$ for accuracy, alongside MAPE and MBE to assess relative errors and biases\footnote{The full methodology and a batch of Dataset will be made publicly available on Github following the peer-review process.}.

\subsection{Preprocessing of the MaRTy Data}

The preprocessing of MaRTy data involved cleaning, normalization, and feature engineering. To ensure consistency, missing values were interpolated using the temporal nearest-neighbor imputation. Outliers, primarily caused by sensor errors, were removed based on a statistical threshold of 3 standard deviations from the mean. 



The original images captured along with MaRTy measurements were six-directional cube maps (North, East, South, West, Up, and Down). To facilitate a more integrated spatial representation, we transformed these cube maps into hemispherical fisheye projections. This transformation was performed using a spherical warping technique, ensuring geometric consistency across directional images \cite{middel2018sky}.

The conversion followed these steps:
\begin{enumerate}
    \item Normalize each directional image to match the same exposure and intensity levels.
    \item Apply an equiangular projection to map cube faces onto a hemispherical surface.
    \item Stitch overlapping regions to create a seamless fisheye representation.
    \item Validate the projection using sky exposure ratios to maintain accuracy.
\end{enumerate}

\begin{figure}[htbp]
  \centering
  \includegraphics[width=\linewidth]{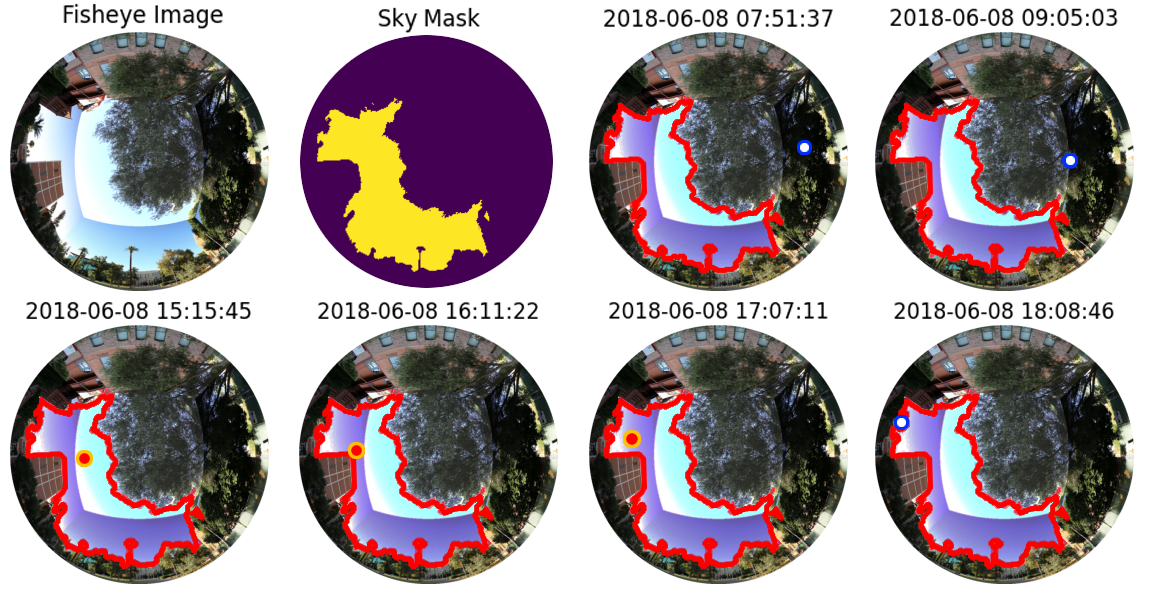}
  \caption{Sky mask generated using SegFormer on the fisheye image, distinguishing sun-exposed and shaded areas based on sun position}
  \label{fig:skymask}
\end{figure}

\noindent
These hemispherical fisheye images are used for extracting shading features, sunlit area estimation, and built environment features enabling a more precise assessment of shortwave and longwave radiation exposure in the model. The precise position of the sun is computed based on geographic metadata, including latitude, longitude, date, and time. To determine whether a location is sun-exposed or shaded, we employ SegFormer to apply a sky mask to the fisheye image, effectively removing the sky pixels and isolating surrounding urban structures and vegetation. This enables us to assess whether the sun is obstructed by nearby objects, providing an image-derived shading estimation without relying on predefined urban morphology datasets (Figure \ref{fig:skymask}).

\subsection{Architecture of the Deep Learning Model}

Our approach involved training two types of neural networks: a standard deep learning model and a Physics-Informed Neural Network (PINN). The standard neural network (NN) served as a baseline, while the PINN incorporated radiative heat transfer physics into its learning process.

\subsubsection{Baseline Neural Network (NN)}

The baseline neural network model was designed as a multimodal fusion network, combining numerical environmental metadata with fisheye-derived spatial features. The architecture included:
\begin{itemize}
    \item Fully connected layers for processing numerical meteorological and built environment features.
    \item Convolutional layers (ResNet-50) for extracting shading features from fisheye images.
    \item A final regression layer predicting $T_{mrt}$.
\end{itemize}

Training was conducted using the Adam optimizer with a learning rate of 0.001, minimizing the Mean Squared Error (MSE) loss function.

\begin{figure*}[htbp]
  \centering
  \includegraphics[width=0.95\linewidth]{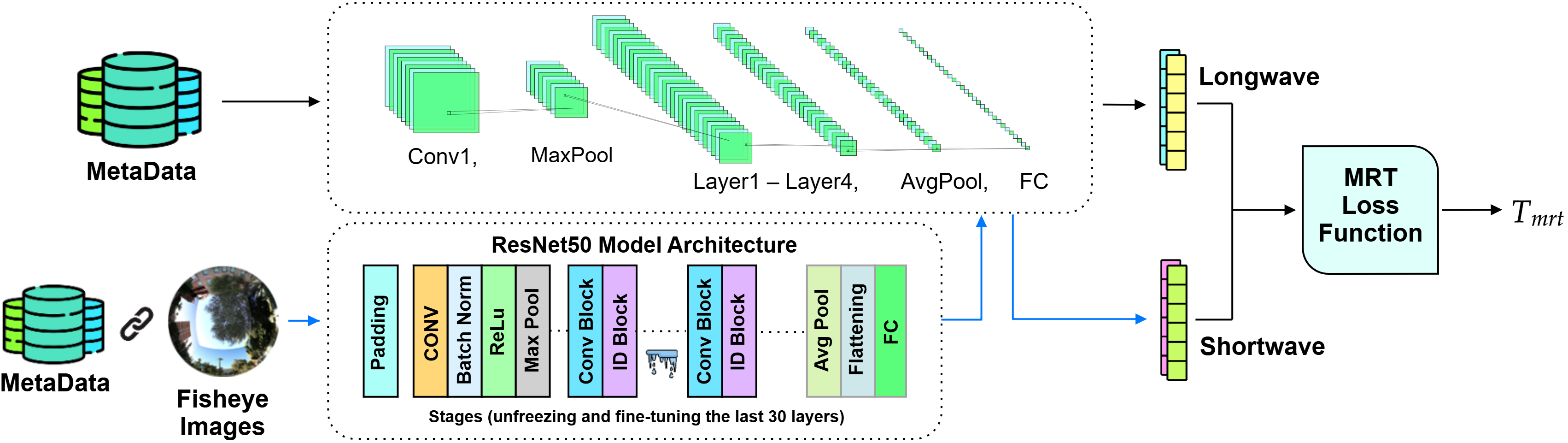}
  \caption{Physics-Informed Neural network architecture for $T_{mrt}$ estimation.}
  \label{fig:nn_architecture}
\end{figure*}

\subsubsection{Physics-Informed Neural Network (PINN)}

To enhance model interpretability and accuracy, we introduced a PINN framework that integrates domain-specific physical constraints. The PINN model incorporated physics-based loss functions derived from the $T_{mrt}$ formula (six directions approach), enforcing consistency with thermodynamic principles. The key radiative heat transfer equation used in the loss function was:
\begin{equation}
    T_{mrt} = \left( \frac{Q_{total}}{a_1 \sigma} \right)^{0.25} - 273.15
\end{equation}
where $Q_{total}$ is the total radiative flux received from different environmental directions, computed as:
\begin{equation}
    Q_{total} = Q_{(up,\ down)} + Q_{(others)}
\end{equation}
The directional flux components are given by:
\begin{equation}
    \begin{aligned}
        Q_{(up,\ down)} = &W_{(up,\ down)} \sum_{i} (a_k S_i + a_l L_i) \\[-0.1cm]
        &\text{\scriptsize \(i \in \{up, down\}\)}
    \end{aligned}
\end{equation}
\begin{equation}
    \begin{aligned}
        Q_{(others)} &= W_{(others)} \sum_{j} (a_k S_j + a_l L_j) \\[-0.2cm]
        &\text{\scriptsize \( j \in \{north, east, south, west\} \)}
    \end{aligned}
\end{equation}

where $S_i$ and $S_j$ represent shortwave radiation components from different directional sources.
$L_i$ and $L_j$ represent longwave radiation components from corresponding directions. $a_k$ and $a_l$ are absorption coefficients for shortwave and longwave radiation, respectively. $W_{(up,\ down)}$ and $W_{(others)}$ are weighting factors representing the fraction of radiative exposure from the respective directions. $a_1$ is the mean radiant absorptivity of the human body. $\sigma$ is the Stefan-Boltzmann constant.

The physics-informed loss function penalizes deviations from these equations, ensuring that the model adheres to fundamental radiation balance principles.





\subsection{Multimodal PINN Architecture}

The multimodal PINN architecture consists of a ResNet-50 backbone for image feature extraction, where the last 30 layers were unfrozen and fine-tuned to capture radiative exposure and shading features for each directional component. Additionally, fisheye images are padded to ensure that spatial features align effectively with solar angles, allowing the model to better distinguish regions of sun exposure and shading in relation to the computed solar position. It also includes two separate multilayer perceptrons (MLPs) for predicting shortwave and longwave radiation. The Shortwave Radiation MLP processes concatenated image features (2048 dimensions) and metadata features, consisting of three hidden layers with 128, 256, and 128 neurons. The Longwave Radiation MLP processes only numerical metadata, consisting of three hidden layers with 128, 256, and 128 neurons. A multimodal fusion layer combining numerical and image-based features to ensure comprehensive radiative flux estimation.
A final output layer predicting six-directional shortwave and longwave radiation components, which are used to compute $T_{mrt}$.


To optimize model performance, we employed Neural Architecture Search (NAS) to explore different network configurations and conducted a random search for hyperparameter tuning. The training process was further refined using k=3 fold cross-validation to ensure robustness and prevent overfitting. A schematic representation of the architecture is presented in Figure \ref{fig:nn_architecture}. The training dataset was split into 80\% for training and 20\% for validation.


\subsection{Evaluation Metrics of the Prediction Model}
Model performance was assessed using four key metrics. The Root Mean Squared Error (RMSE) quantifies the average prediction error magnitude, with lower values indicating better accuracy:
\begin{equation}
    \text{RMSE} = \sqrt{\frac{1}{n} \sum_{i=1}^{n} (y_i - \hat{y}_i)^2}
\end{equation}

The Coefficient of Determination ($R^2$) measures how well the model explains variance in $T_{mrt}$, with values closer to 1 signifying better predictions:
\begin{equation}
    R^2 = 1 - \frac{\sum_{i=1}^{n} (y_i - \hat{y}_i)^2}{\sum_{i=1}^{n} (y_i - \bar{y})^2}
\end{equation}
The Mean Absolute Percentage Error (MAPE) expresses the relative prediction error as a percentage:
\begin{equation}
    \text{MAPE} = \frac{100}{n} \sum_{i=1}^{n} \left| \frac{y_i - \hat{y}_i}{y_i} \right|
\end{equation}
Lastly, the Mean Bias Error (MBE) identifies systematic over- or underestimation in predictions:
\begin{equation}
    \text{MBE} = \frac{1}{n} \sum_{i=1}^{n} (y_i - \hat{y}_i)
\end{equation}
where $y_i$ represents actual values, $\hat{y}_i$ are predictions, $\bar{y}$ is the mean of actual values, and $n$ is the total number of samples. These metrics ensure a comprehensive evaluation of model accuracy, reliability, and bias.

\section{Results}
\label{sec:results}

\subsection{Shade Prediction Accuracy}
A crucial component of our model is the fisheye-derived shade prediction. To validate shade accuracy, we compared the predicted shade feature with the ground truth shade measured during field data collection. The results show an overall accuracy of 94\%, indicating that our method effectively distinguishes sun-exposed and shaded areas.
\begin{align*}
    \text{Accuracy} = \frac{\text{Correctly Predicted Shade}}{\text{Total Instances}} \times 100\% = \textbf{94\%}
\end{align*}

This high accuracy suggests that our sky mask and sun position calculations reliably estimate shading conditions, reinforcing the effectiveness of fisheye images for radiation modeling.


\subsection{Machine Learning Model Performance}
To establish a strong baseline, we evaluated traditional machine learning methods and tree-based models on MaRTy data for the instances with available site images using the WebMRT~\cite{alkhaled2024webmrt} feature set, including: air temperature, relative humidity, and wind speed, altitude above sea level and presence of shade at measurement, and built environment prperties such as sky view factor and percentages of surrounding trees, buildings, and impervious surfaces, and extracted features including time in minutes from sunrise, sun azimuth angle, and sun altitude angle. These models performed exceptionally well, with XGBoost achieving the lowest RMSE. The results are presented in Table~\ref{tab:ml_performance}.

\vspace{0.1cm}\noindent
\textbf{Experimental Setup}: Our experiments were conducted on a high-performance computing system with:
\begin{itemize}
    \item \textbf{GPU:} NVIDIA Tesla T4 (16 GB VRAM).
    \item \textbf{RAM:} 16 GB system memory.
    \item \textbf{Processor:} Intel Xeon CPU.
    \item \textbf{Operating System:} Linux-based environment.
\end{itemize}

\begin{table}[htbp]
    \centering
    \scriptsize
    \begin{tabular}{lcc}
        \toprule
        Model & RMSE & $R^2$ \\
        \midrule
        Ridge Regression & 5.48 & 0.771 \\
        Lasso Regression & 5.56 & 0.762 \\
        \midrule
        SVR & 4.01 & 0.875 \\
        \midrule
        Decision Tree & 4.72 & 0.821 \\
        Random Forest & 3.49 & 0.889 \\
        LightGBM & 3.48 & 0.891 \\
        XGBoost & 3.48 & 0.893 \\
        \bottomrule
    \end{tabular}
    \caption{Performance of machine learning models on the WebMRT feature set.}
    \label{tab:ml_performance}
\end{table}

\subsection{Ablation Study: Impact of Features and Images}
To analyze the contributions of different input features, we conducted an ablation study by testing various configurations of neural networks (NNs) and PINNs. We examined models trained with the full WebMRT feature set as a baseline, as well as variations where built environment features such as sky view factor (SVF), surrounding buildings, and trees were excluded. Additionally, we evaluated the impact of incorporating fisheye-derived predicted shade features and features extracted from fisheye images using ResNet-50. Finally, we compared the performance of standard NNs and PINNs to assess the benefits of embedding physical constraints into the learning process.

The results in Table~\ref{tab:dl_performance} show that models using the full WebMRT feature set performed consistently well. In the absence of rich metadata, the best-performing PINN model with fisheye images achieved an RMSE of 3.50, demonstrating its effectiveness. Additionally, models using predicted shade instead of measured shade performed similarly, confirming the accuracy of our fisheye-based shading estimation. These findings underscore the adaptability of our approach, especially when conventional input features are incomplete or unavailable.

\begin{table}[htbp]
    \centering
    \scriptsize
    \renewcommand{\arraystretch}{1.1}
    \setlength{\tabcolsep}{3pt}
    \begin{tabular}{lcccc}
        \toprule
        Model & RMSE & $R^2$ & Features & Images \\
        \midrule
        Neural Network & 3.65 & 0.87 & WebMRT Set & \xmark \\
        Neural Network & 6.33 & 0.63 & Shade \xmark, Built \cmark & \xmark \\
        Neural Network & 4.25 & 0.81 & Pred. Shade \cmark, Shade \xmark, Built \xmark & \xmark \\
        Neural Network & 10.58 & 0.32 & Shade \xmark, Built \xmark & \xmark \\
        Multimodal NN & 6.06 & 0.66 & Shade \xmark, Built \xmark & \cmark \\
        Multimodal NN & 3.87 & 0.86 & Pred. Shade \cmark, Shade \xmark, Built \xmark & \cmark \\
        \midrule
        PINN & 3.63 & 0.88 & WebMRT Set & \xmark \\
        PINN & 6.02 & 0.65 & Shade \xmark, Built \cmark & \xmark \\
        PINN & 9.80 & 0.34 & Shade \xmark, Built \xmark & \xmark \\
        PINN & 3.67 & 0.88 & Pred. Shade \cmark, Shade \xmark, Built \xmark & \xmark \\
        \textbf{Multimodal PINN} & \textbf{3.46} & \textbf{0.89} & \textbf{WebMRT Set} & \cmark \\
        \textbf{Multimodal PINN} & \textbf{3.54} & \textbf{0.87} & \textbf{Shade} \xmark, \textbf{Built} \xmark & \cmark \\
        \textbf{Multimodal PINN} & \textbf{3.50} & \textbf{0.88} & \textbf{Pred. Shade} \cmark, \textbf{Shade} \xmark, \textbf{Built} \xmark & \cmark \\
        \bottomrule
    \end{tabular}
    \caption{Performance of deep learning models across different feature sets.}
    \label{tab:dl_performance}
\end{table}

\noindent
Table \ref{tab:dl_performance} shows that fisheye-derived shading enables PINN models to achieve WebMRT-level accuracy (RMSE = 3.50 vs. 3.48). This confirms that $T_{mrt}$ can be estimated without reliance on the measured urban metadata, improving adaptability.

To further analyze performance, we compare key models based on RMSE, $R^2$, MAPE, and MBE (Table \ref{tab:comparison_performance}).

\begin{table}[htbp]
    \centering
    \scriptsize
    \renewcommand{\arraystretch}{1.1}
    \setlength{\tabcolsep}{3pt}
    \begin{tabular}{lcccc}
        \toprule
        Model & RMSE & $R^2$ & MAPE & MBE \\
        \midrule
        Neural Network & 3.65 & 0.879 & 6.2 & 0.24 \\
        Multimodal Neural Network & 3.87 & 0.863 & 7.47 & 0.31 \\
        Physics-Informed Neural Network & 3.63 & 0.881 & 6.1 & 0.22 \\
        Multimodal Physics-Informed Neural Network & 3.50 & 0.889 & 6.3 & 0.26 \\
        \bottomrule
    \end{tabular}
    \caption{Comparison of selected models using RMSE, $R^2$, MAPE, and MBE metrics.}
    \label{tab:comparison_performance}
\end{table}

This comparison highlights that the PINN with fisheye-based features and predicted shading achieves nearly the same performance as the full WebMRT feature set while eliminating the need for manually provided built environment attributes. These results validate the effectiveness of our approach in estimating $T_{mrt}$ using image-derived urban features, making it scalable for large-scale applications where detailed metadata may not be available.

\begin{figure}[ht]
    \centering
    \includegraphics[width=\linewidth]{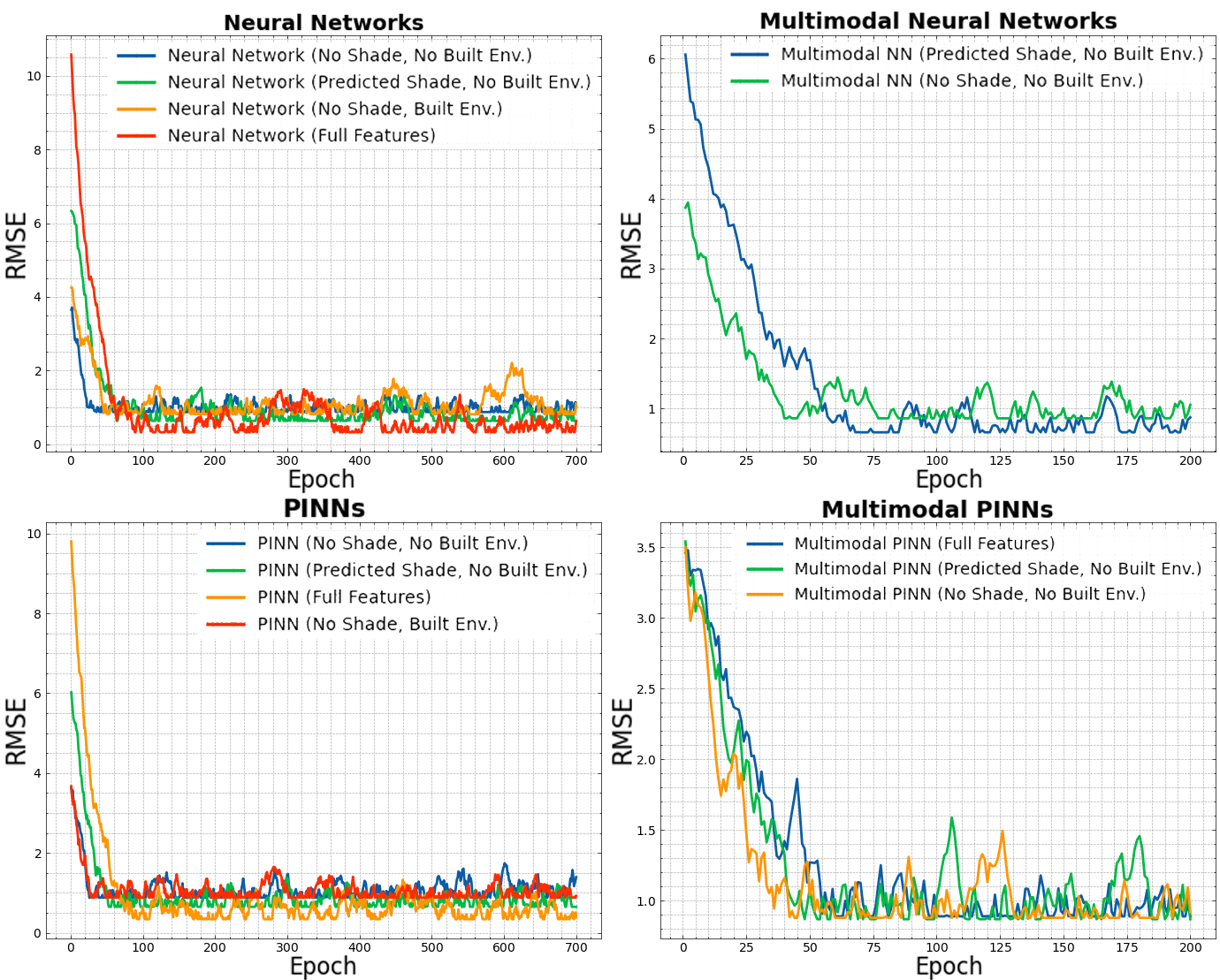}
    \caption{Comparison of RMSE for different machine learning and deep learning models.}
    \label{fig:rmse_comparison}
\end{figure}

The comparative performance of different models is visually depicted in Figure~\ref{fig:rmse_comparison}, illustrating RMSE variations across different configurations.

\section{Discussion}
\label{sec:discussion}

Our findings demonstrate that incorporating PINNs into $T_{mrt}$ estimation enhances both accuracy and physical consistency compared to data-driven models like WebMRT~\cite{alkhaled2024webmrt}. Unlike empirical models such as ENVI-met and SOLWEIG~\cite{sinsel2022implementation}, which require extensive computational resources, our PINN framework offers a transferable intelligence that remains robust even when built environment metadata and predifined shading info is unavailable. Additionally, unlike the GA-BPNN model proposed by Xie et al.~\cite{xie2022backpropagation}, which relies on statistical feature extraction, our approach directly integrates radiative transfer equations and extracted features, ensuring thermodynamic consistency.

Our method dynamically infers shading from fisheye imagery, achieving 94\% agreement with measured data, proving its reliability as a field measurement alternative. Compared to previous $T_{mrt}$ estimation models, such as those developed by Jia et al.~\cite{jia2024estimation} and Xie et al.~\cite{xie2022backpropagation}, which often struggle with dataset biases, our approach combines computer vision and physics-informed learning to improve adaptability across diverse urban settings. Furthermore, our work extends prior studies like ``Fifty Grades of Shade" by Middel et al.~\cite{middel202150}, which emphasized the role of shading in pedestrian thermal comfort, by demonstrating that fisheye-derived features can accurately estimate shading conditions without requiring direct field observations.

The following discussion revisits our research questions, highlighting how our framework improves accuracy, shading estimation, and generalizability while maintaining physical consistency.

\noindent
\textbf{Answer to \ref{rq1}:} Our PINN framework, particularly when incorporating fisheye-derived shading, demonstrated superior performance in scenarios with incomplete metadata. It provided results comparable to traditional empirical models when sufficient environmental data was available while maintaining physical consistency through radiative balance constraints.

\noindent
\textbf{Answer to \ref{rq2}:} Our approach achieved a 94\% agreement between predicted shading and measured shading conditions, validating the accuracy of fisheye-based shading estimation. By integrating these image-derived features, our model captured dynamic shading effects without relying on predefined urban morphology datasets, significantly improving $T_{mrt}$ prediction.

\noindent
\textbf{Answer to \ref{rq3}:} Our model performed well in desert urban environments, where shading and solar radiation play a crucial role in thermal comfort. However, its applicability to temperate and humid climates requires additional validation, as urban materials, vegetation density, and atmospheric conditions differ significantly across regions.

\noindent
\textbf{Answer to \ref{rq4}:} The PINN framework enforces radiative energy balance constraints, ensuring that $T_{mrt}$ estimates adhere to fundamental thermodynamic principles. This prevents physically unrealistic predictions commonly found in purely data-driven models and improves interpretability.

\noindent
\textbf{Answer to \ref{rq5}:} By leveraging vision-based feature extraction, our framework reduces dependency on costly and labor-intensive radiation sensors. Our results show that PINN with fisheye-derived shading features achieves accuracy comparable to models using full environmental metadata, suggesting a viable alternative for large-scale, sensor-independent thermal assessments.

\vspace{0.2cm}
\noindent
\textbf{Limitations:} Despite promising results, our approach has several limitations. The dataset was collected in desert regions, meaning the model may not generalize well to colder climates without additional data. Additionally, the multimodal PINN framework requires significant GPU resources, which may limit real-time deployment. Finally, further validation in diverse urban environments with varied architectural and vegetation layouts is necessary to enhance generalizability.
\section{Conclusion}
\label{sec:conclusion}

In this study, we introduced a novel PINN framework for estimating $T_{mrt}$, integrating radiative transfer principles with deep learning. Our approach bridges the gap between traditional physics-based models and data-driven machine learning methods by leveraging multimodal environmental data, including fisheye images and metadata. The results indicate that our method successfully addresses key challenges in $T_{mrt}$ estimation, particularly in urban environments where radiation dynamics are highly complex.

Our framework demonstrates that PINNs, when linked with fisheye-derived features, can achieve high accuracy even in scenarios where complete environmental metadata is unavailable. The PINN model proved to be the superior choice when direct shading or surrounding building data was missing. The best-performing configuration, which included fisheye-derived shading and surrounding information, achieved an RMSE of 3.50, outperforming alternative approaches in data-limited scenarios. This confirms that fisheye images serve as a reliable proxy for capturing local shading conditions, allowing the model to infer radiative flux variations without requiring explicit metadata input. 

\vspace{0.2cm}
\noindent
\textbf{Future Work}

Future work will focus on optimizing the PINN architecture to improve computational efficiency and enable six-direction image input. Additionally, expanding the dataset to include urban environments with varying climatic conditions will help validate the model's scalability. Investigating alternative deep learning architectures, such as transformer-based models, will enhance spatial feature extraction from fisheye images. Lastly, integrating this framework into existing urban climate simulation tools will provide a more comprehensive approach to assessing outdoor human thermal exposure in cities.

By refining and extending our approach, this research lays the foundation for a more scalable and interpretable $T_{mrt}$ estimation framework, ultimately aiding in climate-adaptive urban planning and public health strategies.

{
    \small
    \bibliographystyle{ieeenat_fullname}
    \bibliography{main}
}

\end{document}